%% file: 0_main.tex
\documentclass[journal]{IEEEtran}

\ifCLASSINFOpdf
\else
   \usepackage[dvips]{graphicx}
\fi
\usepackage{url}

\usepackage{graphicx}
\usepackage{amsfonts,amsmath}
\usepackage[pagebackref=false,breaklinks=true,colorlinks,citecolor=black,linkcolor=black,bookmarks=false]{hyperref} 
\usepackage{float}
\usepackage{graphicx}
\usepackage{subfig}
\usepackage{amssymb}
\usepackage{multirow}

\begin{document}

\title{Dynamic Point Cloud Geometry Compression Using Multiscale Inter Conditional Coding}

\author{Jianqiang Wang, Dandan Ding, Hao Chen, and Zhan Ma
\thanks{J.~Wang, H. Chen and Z.~Ma are with Nanjing University, China; D.~Ding is  with Hangzhou Normal University, China.}}

\maketitle

\begin{abstract}
This work extends the Multiscale Sparse Representation (MSR) framework developed for static Point Cloud Geometry  Compression (PCGC) to support the dynamic PCGC through the use of multiscale inter conditional coding. To this end, the reconstruction of the preceding Point Cloud Geometry (PCG) frame is progressively downscaled  to generate multiscale temporal priors which are then scale-wise transferred and integrated with lower-scale spatial priors from the same frame to form the contextual information to  improve occupancy probability approximation when processing the current PCG frame from one scale to another.   
Following the Common Test Conditions (CTC) defined in the standardization committee, the proposed method presents State-Of-The-Art (SOTA) compression performance, yielding 78\% lossy BD-Rate gain to the latest standard-compliant V-PCC and  45\% lossless bitrate reduction to the latest G-PCC. Even for recently-emerged learning-based solutions, our method still shows significant performance gains.
\end{abstract}


\IEEEpeerreviewmaketitle

\input{1_intro}
\input{2_related}
\input{3_method}

\input{4_exp}
\input{6_conclusion}
\newpage
\bibliographystyle{IEEEbib}
\bibliography{refs}

\end{document}

%% file: 1_intro.tex
\section{Introduction}

\begin{figure*}[thbp]
\centering
	\subfloat[]{\includegraphics[height=1.75in]{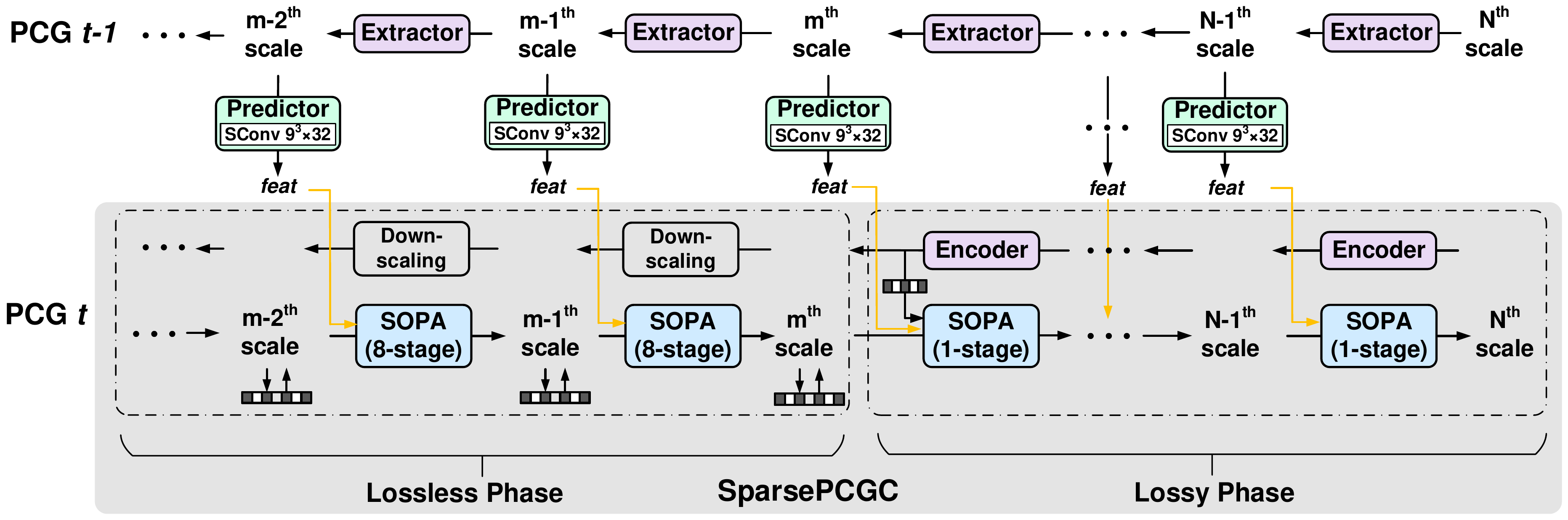}}
    \quad
	\subfloat[]{\includegraphics[height=1.75in]{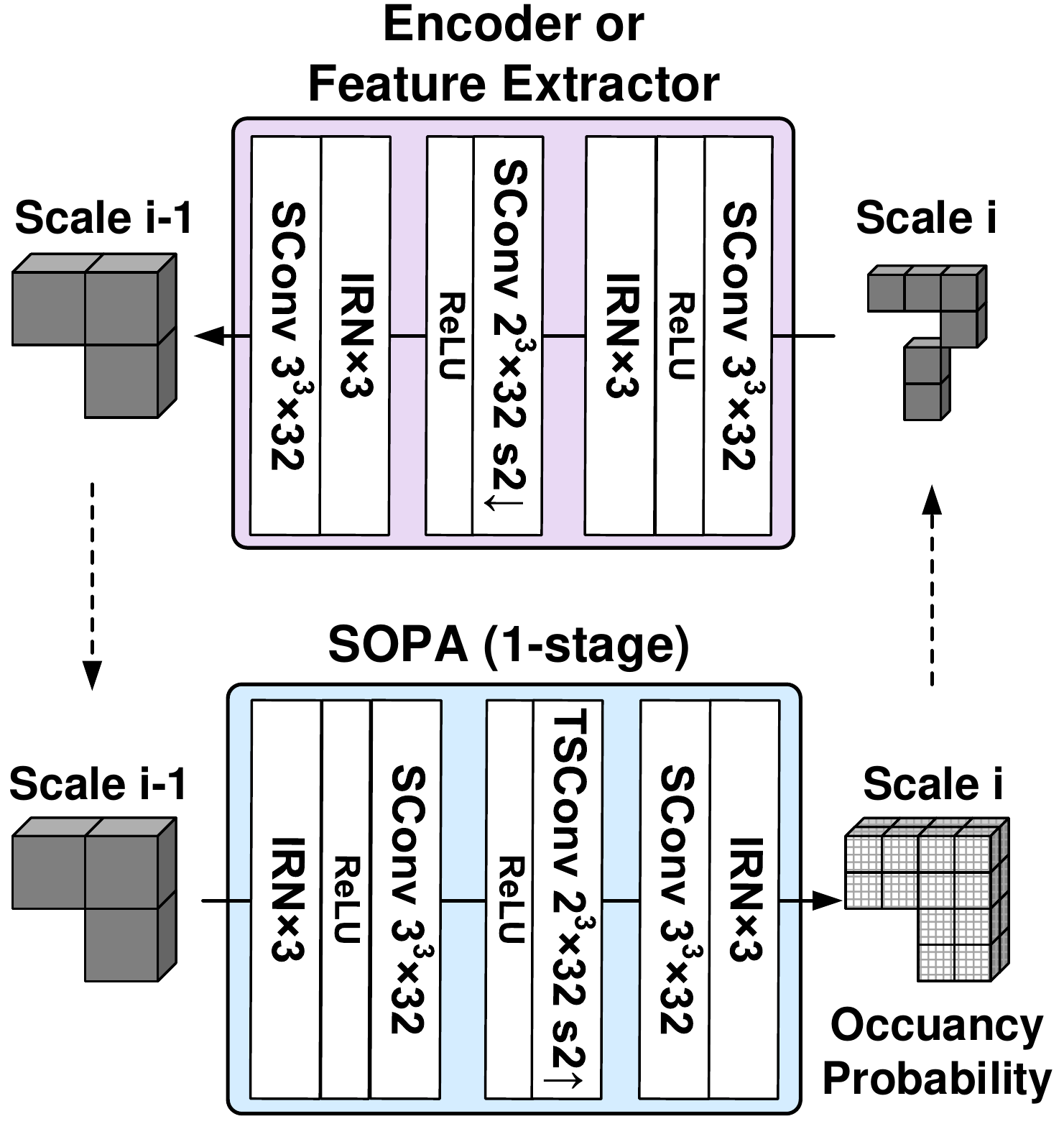}\label{fig:network}}
	\caption{{\bf Dynamic PCGC in a two-frame example.} (a) On top of the MSR framework used by SparsePCGC for static PCGC originally, multiscale temporal priors of ($t-1$)-th frame are first extracted using Extractors and transferred using Predictors for  the compression of $t$-th frame, where temporal priors are concatenated with the same-frame lower-scale priors for improving the capacity of SOPA model. (b) Network examples for Encoder (or Feature Extractor) and 1-stage SOPA.  {Lossy SparsePCGC is comprised of a lossless phase using 8-stage SOPA and a lossy phase using 1-stage SOPA instead, across different scales. On the contrary, lossless SparsePCGC uses 8-stage SOPA for all scales~\cite{Wang2021SparseTM}.}  Sparse Convolution (SConv) constitutes the basic feature processing layer.  Inception-ResNet (IRN) blocks are used for deep feature aggregation~\cite{Wang2021MultiscalePC}.}
	\label{fig:overview}
\end{figure*}

Dynamic point clouds are of great importance for applications like holographic communication, autonomous machinery, etc., for which the efficient compression of dynamic Point Cloud Geometry (PCG) plays a vital role in  service provisioning. In addition to rules-based Point Cloud Geometry Compression (PCGC) technologies standardized by the ISO/IEC MPEG (Moving Picture Experts Group), e.g.,  Video-based PCC (V-PCC) and Geometry-based PCC (G-PCC)~\cite{graziosi2020overview,schwarz2018emerging,cao2021compression}, learning-based PCGC methods have been extensively investigated in the past few years, greatly improving the performance with very encouraging prospects~\cite{quach2022survey}. Among those learning-based solutions, multiscale sparse representation (MSR)~\cite{Wang2021MultiscalePC, Wang2021SparseTM,PCGFormer, Xue2022EfficientLP} has improved the performance unprecedentedly by effectively exploiting cross-scale and same-scale correlations in the same frame of a static PCG for compact representation.
The compression of a static PCG frame independently  can also be referred to as the intra coding of the PCG.

This work extends the MSR framework originally developed for static PCGs to compress the dynamic PCGs~\cite{Wang2021MultiscalePC,Wang2021SparseTM}. In this regard, we suggest the inclusion of multiscale temporal priors for inter conditional coding. As in Fig.~\ref{fig:overview}, for a previously-reconstructed PCG frame (e.g., PCG $t-1$), we progressively downsample it and extract scale-wise hierarchical features which are then transferred as additional temporal priors to help the compression of the same-scale PCG tensor of the current frame (e.g., PCG $t)$. To this end, we basically concatenate the same-scale temporal priors from the inter reference and lower-scale spatial priors from the same intra frame to form the contextual information for better conditional occupancy probability approximation in compression.  Such an inter conditional coding scheme for dynamic PCGC is implemented on top of the SparsePCGC~\cite{Wang2021SparseTM}  originally developed for the static PCGC, to quantitatively evaluate its efficiency. 
Experimental results demonstrate the leading  performance of our method when compared with existing methods (either rules-based or learning-based ones) in both lossy and lossless modes, following the Common Test Conditions (CTC) used in the MPEG standardization committee~\cite{MPEG-EE-AIDPCC}. 

%% file: 2_related.tex
\section{Related Work}
In addition to existing G-PCC and V-PCC standards and other rules-based PCC methods in~\cite{Intra_PCG,PCAC_Graph,Silhouette4D_w_Context,7405340,8784388,gu20193d}, an excessive number of learning-based PCC solutions have emerged in the past years. Therefore the ISO/IEC MPEG 3D graphics  coding group initiated the Artificial Intelligence-based Point Cloud Compression (AI-PCC) to investigate potential technologies for better compression of point clouds.

{{\bf Static PCGC}.} Recently, major endeavors have been paid to study the  compression of a static PCG~\cite{quach2022survey}, a.k.a. Static PCGC, yielding voxel-based~\cite{Wang2021LossyPC,Voxel_PCGC}, point-based~\cite{Zhang_Transformer_PCGC}, octree-based~\cite{Huang2020OctSqueezeOE}, and sparse tensor-based approaches~\cite{PCGFormer,Wang2021MultiscalePC,Wang2021SparseTM}.
Among them, sparse tensor-based methods not only attain the leading performance but also have low complexity. The first representative work is the PCGCv2~\cite{Wang2021MultiscalePC} where a static PCG tensor is hierarchically downsampled and lossily compressed using a Sparse Convolutional Neural Network (SparseCNN) based autoencoder. Later,  the SparsePCGC~\cite{Wang2021SparseTM} improves the PCGCv2 greatly under a unified MSR framework to support both lossless and lossy compression of various point clouds by extensively exploiting cross-scale and same-scale correlations for better contextual modeling using SparseCNN-based Occupancy Probability Approximation (SOPA) models.
More details regarding the MSR and SOPA model can be found in~\cite{Wang2021SparseTM}.

{\bf Dynamic PCGC.} On top of the PCGCv2, Fan~\textit{et al.}~\cite{Fan2022DDPCCDD} and Akhtar~\textit{et al.}~\cite{Akhtar2022InterFrameCF} proposed to encode inter residuals between temporal successive PCG frames for dynamic PCGC. Their main difference lies in the generation of inter prediction signals.
Fan~\textit{et al.}~\cite{Fan2022DDPCCDD} used a SparseCNN-based motion estimation to align the coordinate of the reference to the current frame, and then interpolate $k$ nearest neighbors to {{first derive the temporal prediction and then compute the residual difference}}; while Akhtar~\textit{et al.}~\cite{Akhtar2022InterFrameCF} employed a ``convolution on target coordinates'' operation to map the feature-space information from the reference to the current frame to derive the inter residual.

This letter also applies the ``convolution on target coordinates'' to exploit correlations across temporal successive frames in feature space. Instead of using the inter residual at a fixed scale, we generate multiscale temporal priors for scale-wise contextual information aggregation, which greatly improves the conditional probability approximation in compression of our method, as shown subsequently.

%% file: 3_method.tex
\section{Proposed Method}

\subsection{Overall Framework}
The proposed MSR-based dynamic PCGC is shown in Fig.~\ref{fig:overview}. A two-frame example is illustrated where the $(t-1)$-th frame is already encoded and reconstructed as the temporal reference, and the $t$-th frame is about to be encoded. Apparently, such a two-frame example can be easily extended to  a sequence of frames.

To compress the $t$-th frame, a straightforward solution is to encode each PCG frame independently, a.k.a. intra coding, using default SparsePCGC to solely exploit cross-scale and same-scale correlations in the same frame. As there are strong temporal correlations across successive frames, inter prediction is often utilized for improving compression efficiency. To this end, this work follows the MSR principle to first progressively extract features using Extractors from the $(t-1)$-th  reconstruction $\hat{x}_{t-1}$, and then generate multiscale temporal priors via a one-layer sparse convolution (SConv) based Predictors for inter conditional coding of $t$-th frame $x_t$.

Similar to the SparsePCGC, dyadic resampling is applied for multiscale computation~\cite{Wang2021SparseTM}. Assuming the highest scale of an input point cloud at $N$, the lossy compression of this PCG is comprised of $m$-scale lossless   and  ($N-m$)-scale lossy compression. Adapting $m$ is to balance the lossy rate-distortion tradeoff~\cite{sullivan1998rate}. As seen, in the lossless phase, temporal priors from the inter reference are concatenated with the lower-scale spatial priors in the same intra frame which are then fed into the 8-stage SOPA model for better approximation of occupancy probability for lossless coding; while in the lossy phase, such concatenated spatiotemporal priors can be either augmented with decoded local neighborhood information or directly used in 1-stage SOPA model for better geometry reconstruction.
By contrast, the lossless compression of an input PCG applies 8-stage SOPA uniformly for all scales to process such concatenated spatiotemporal priors.

We next detail each individual module developed for the use of multiscale temporal priors in inter conditional coding.

\subsection{Encoder/Extractor \& SOPA Models}

The Encoder (and Extractor) model  which is typically devised with the resolution downscaling, aggregates  local neighborhood information to form spatial intra (or temporal inter) priors for enhancing the SOPA model. Correspondingly, the SOPA model estimates the occupancy probability for geometry reconstruction (i.e., voxel occupancy status) gradually from lower to higher scale, using both spatial priors (e.g., decoded latent feature, lower-scale input) in the same frame and temporal priors from the inter reference. 

The Encoder/Extractor model applies sparse convolutions and nonlinear activations for computation as shown in Fig.~\ref{fig:network}, consisting of a convolutional voxel downsampling layer with kernel size and stride of 2 at each dimension, e.g., SConv $2^3\times32$ s2$\downarrow$, and stacked Inception-ResNet (IRN) blocks for deep  feature aggregation. The IRN contains multiple convolutional layers with a kernel size of 3$\times$3$\times$3, e.g., SConv $3^3\times32$~\cite{Wang2021MultiscalePC}.

{The 1-stage SOPA model mostly mirrors the processing of the Encoder/Extractor where a transposed convolutional voxel upsampling layer with kernel size and stride of 2 is used, e.g., SConv $2^3\times32$ s2$\uparrow$. This 1-stage SOPA can be easily extended to support multi-stage computation by grouping upscaled voxels for stage-wise processing~\cite{Wang2021SparseTM}. As exemplified in the lossless phase of Fig.~\ref{fig:overview}, 8-stage SOPA is used to progressively reconstruct the voxels by utilizing previously-processed, same-scale neighbors for better probability estimation. }

\begin{figure}[t]
\centering
\subfloat[]{\includegraphics[height=1.10in]
{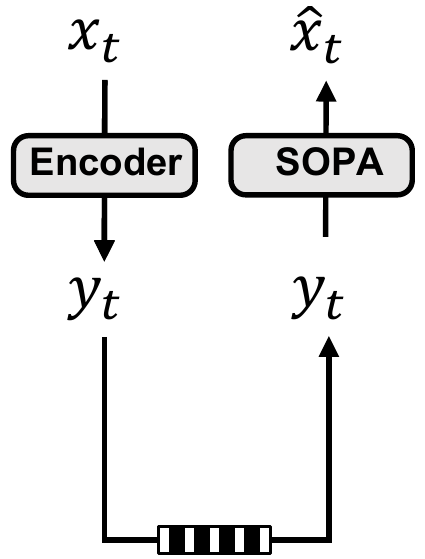}\label{sfig:intra_coding}}
\subfloat[]{\includegraphics[height=1.10in]{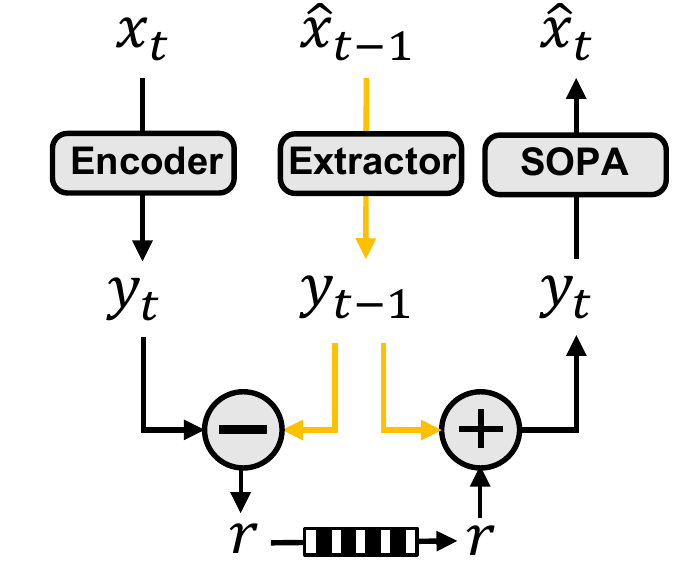}\label{sfig:inter_residual_coding}}
\subfloat[]{\includegraphics[height=1.10in]{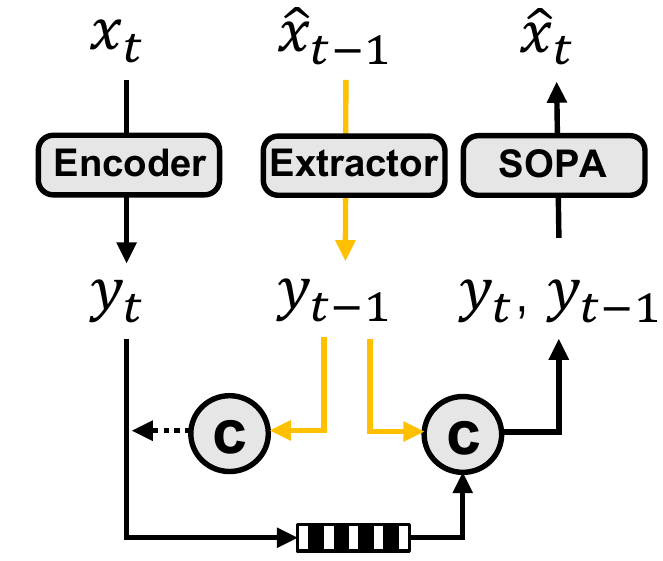} \label{sfig:inter_conditional_coding}}
\caption{{\bf Intra and Inter Coding of PCGs.} (a) intra coding used in SparsePCGC~\cite{Wang2021SparseTM}, (b) inter residual coding used in~\cite{Fan2022DDPCCDD,Akhtar2022InterFrameCF}, (3) the proposed inter conditional coding.}
\label{fig:compensation}
\end{figure}

\subsection{Inter Conditional Coding}
We use the Predictor to transfer information from the inter reference for the compression of  the current frame. 
As in Fig.~\ref{fig:overview}, the Predictor is implemented using a one-layer sparse convolution to perform the ``convolution on target coordinates'', which has the same number of parameters and operations as the normal convolution, except the target coordinates of its output can be customized. For instance,  
a sparse tensor is formulated using a set of coordinates $\vec{C}=\{(x_i, y_i, z_i)\}_i$ and associated features $\vec{\bf F}=\{\vec{f}_i\}_i$. The sparse convolution is formulated as :
\begin{equation}
\vec{f}_{u}^{out} = \sum\nolimits_{k \in \mathbb{N}^{3}(u, \vec{C}^{in})} W_{k} \vec{f}_{u+k}^{in} 
\quad\text{for}\quad
u \in \vec{C}^{out},
\end{equation} 
where $\vec{C}^{in}$ and $\vec{C}^{out}$ are input and output coordinates in the reference frame and current frame, respectively.
$\mathbb{N}^{3}(u, \vec{C}^{in}) = \{k|u+k \in \vec{C}^{in}, k\in \mathbb{N}^{3}\}$ defines a 3D convolutional kernel centered at $u \in \vec{C}^{out}$ with offset $k$ in $\vec{C}^{in}$. 
$\vec{f}_{u+k}^{in}$ and $\vec{f}_{u}^{out}$ are corresponding input and output feature vectors at coordinate $u+k \in \vec{C}^{in}$ and $u \in \vec{C}^{out}$, respectively. 
$W_i$ is kernel weights.  In this work, the Predictor takes each coordinate of the current frame as the center, aggregates, and transfers the colocated features at each scale in a $9\times9\times9$ local window of the reference, e.g., SConv $9^3\times32$.

The use of temporal priors $y_{t-1}$  from the reference for inter prediction is exemplified in Fig.~\ref{fig:compensation}. As for a comparison, intra coding is also pictured in Fig.~\ref{sfig:intra_coding}. {The inter  residual coding scheme is used in~\cite{Fan2022DDPCCDD, Akhtar2022InterFrameCF} where the feature residual between the reference  $y_{t-1}$ and current frame is encoded as in Fig.~\ref{sfig:inter_residual_coding}}. 
The residual compensation is usually limited at the first layer of the lossy phase because it requires the correct geometry information for augmentation. Having residual compensation in other lossy scales is impractical because incorrect geometry  would severely degrade the reconstruction quality~\cite{Wang2021MultiscalePC}.

By contrast,  a simple-yet-effective spatiotemporal feature concatenation is applied to perform the inter conditional coding in Fig.~\ref{sfig:inter_conditional_coding} which  is  flexible and applicable to all scales under the MSR framework.  As seen, the reference reconstruction $\hat{x}_{t-1}$ is used to generate scale-wise temporal priors which are then concatenated with the (cross-scale) spatial priors from the same frame to help the compression in both lossless and lossy compression.
In this way, we retain all the information of temporal reference and use it for the compression of $y_{t}$, which allows the codec to adaptively extract useful information for occupancy probability estimation. In lossless mode, it generates bitstream with less bitrate consumption; while in lossy mode, it helps to better reconstruct the geometry with less distortion.

\subsection{Loss Functions}
To quantify the voxel occupancy probability, we use the Binary Cross-Entropy (BCE) loss to measure the bitrate required to encode the occupancy status. At the same time, the BCE loss also represents the geometry distortion in lossy compression.
For the compression of latent feature in the encoder, we use a simple factorized entropy model~\cite{balle2018variational} to estimate its probability, and cross-entropy loss to calculate the bitrate $R_{F}$.
The total loss function is the combination of  BCE loss and rate consumption $R_{F}$, i.e., $Loss = \text{BCE} + \lambda \cdot R$, where $\lambda$ is the weight used to adjust the rate-distortion tradeoff.

%% file: 4_exp.tex
\input{figs/table}

\begin{figure}[t]
	\centering
	\includegraphics[width=1.65in]{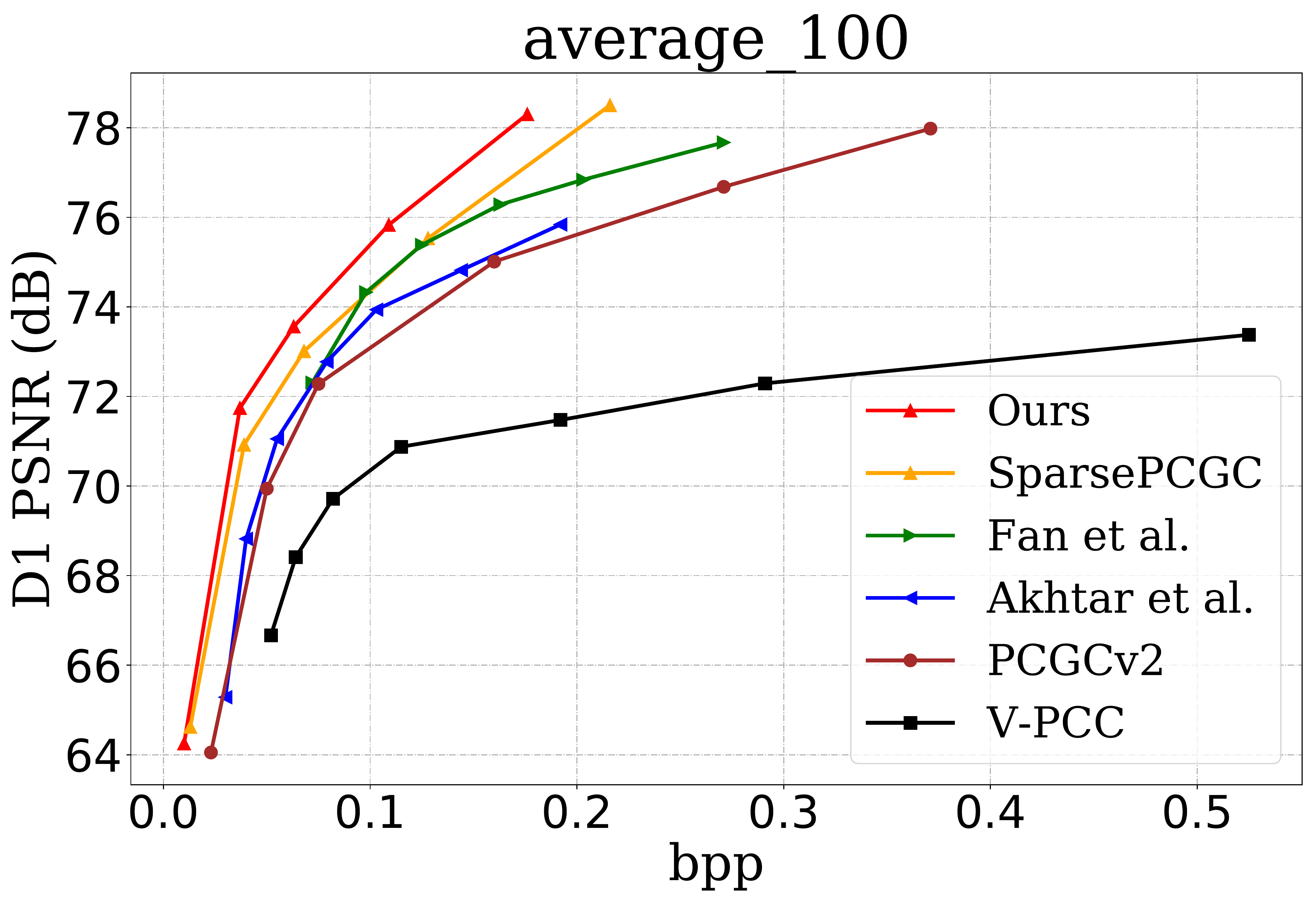}
	\includegraphics[width=1.65in]{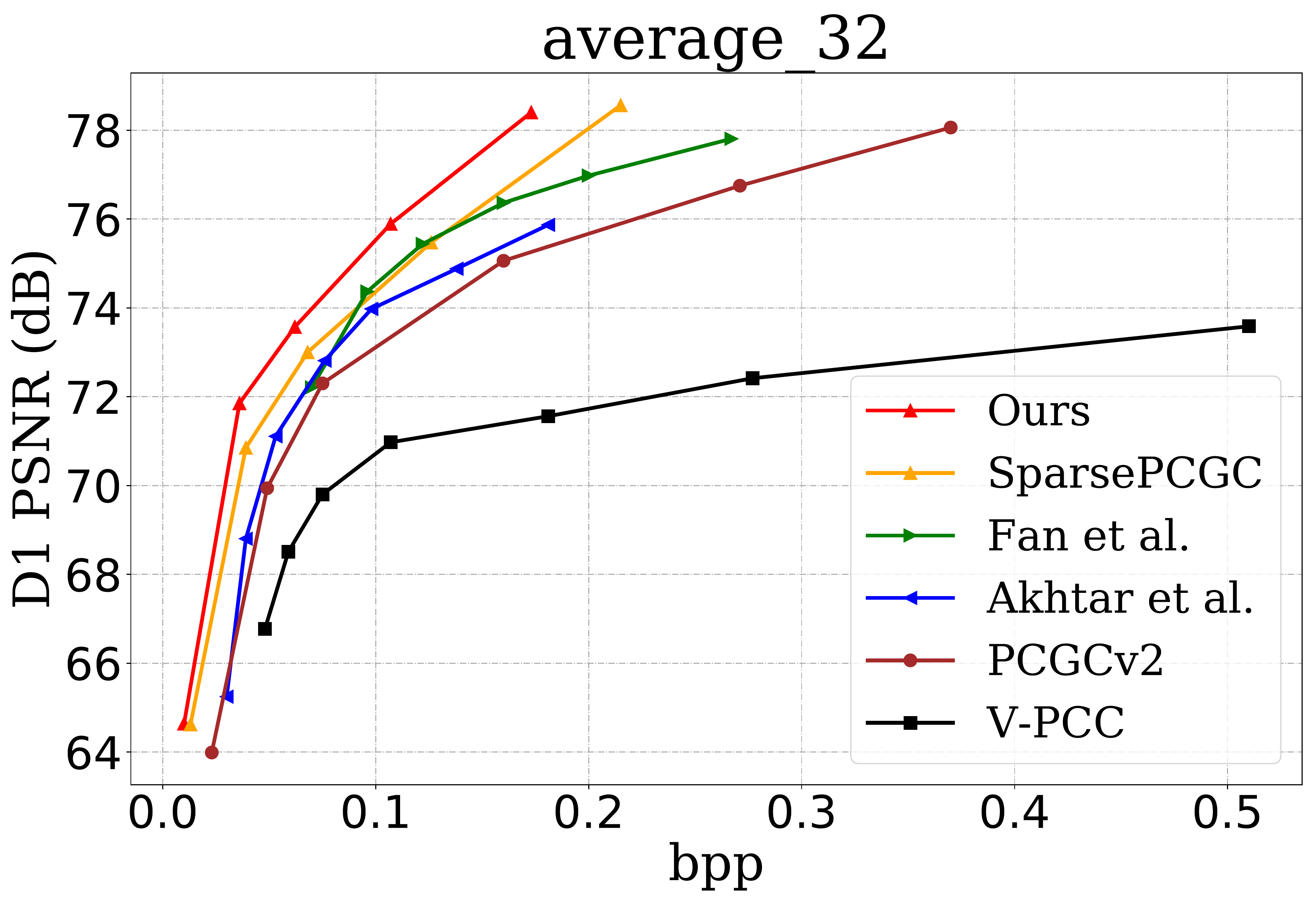}
	\caption{{\bf Efficiency Comparison.} Rate-Distortion (R-D) curves of different methods. 100 (left) and 32 (right) frames are evaluated across a wide range of bitrates following the CTC~\cite{MPEG-EE-AIDPCC}.}
\label{fig:rdcurves}
\end{figure}

\section{Experimental Results}

\subsection{Testing and Training Conditions}
\textbf{Training and Testing Datasets.}
We use the 8i Voxelized Full Bodies (8iVFB) dataset~\cite{8i20178i} for training and the Owlii dynamic human sequence dataset~\cite{xu2017owlii} for testing. The training dataset contains 5 sequences: \textit{longdress}, \textit{loot}, \textit{redandblack}, \textit{soldier}, \textit{queen}, each of which has 300  frames at 10-bit geometry precision. 
The test dataset contains 4 sequences: \textit{basketball\_player}, \textit{dancer}, \textit{model}, \textit{exercise}. They are all quantized to 10-bit geometry precision. 
The splitting of training and testing samples follows the Exploration Experiment (EE) recommendations used in MPEG AI-PCC group~\cite{MPEG-EE-AIDPCC}.  

\textbf{Training Strategies.}
In training, we partition each frame into 4 blocks with kdtree and progressively downscale them to 4 different scales for data augmentation.
We train one model for lossless coding and five models for lossy coding. By adjusting $m$ in lossy phase and the R-D weight $\lambda$ in  the loss function, we obtain five different lossy coding models, covering bitrates from 0.01  to 0.18 bpp (bits per point).

\textbf{Testing Conditions.}
The testing follows the common test condition (CTC) defined in the AI-PCC group for  dynamic PCGC~\cite{MPEG-EE-AIDPCC}. The first frame is encoded in intra mode, followed by all P frames that use the temporally-closest reconstruction as the reference.
Results are averaged for cases using 32 frames and {100} frames.
The bitrate is evaluated by the average bits per input point (bpp) for each sequence. The geometric distortion is evaluated by D1-PSNR per frame to produce a sequence-level average (the first intra frame is also included).

\subsection{Performance Evaluation}

For lossy coding, the V-PCC~\cite{vpcc2021standard} is selected for comparison because of its SOTA performance for dynamic lossy PCGC; Here we apply the default low-delay HEVC video encoding in V-PCC. While for lossless coding, the G-PCC (octree) is compared because of its superior efficiency.
Moreover, we compare with other learning-based PCGC methods, including the PCGCv2~\cite{Wang2021MultiscalePC} and the SparsePCGC~\cite{Wang2021SparseTM} which were originally developed for static PCGC, and two recently-emerged dynamic PCGC methods proposed by Akhtar {\it et al.}~\cite{Akhtar2022InterFrameCF} and Fan {\it et al.}~\cite{Fan2022DDPCCDD}. For the PCGCv2 and SparsePCGC, every PCG frame is coded independently as intra mode without inter prediction.
Regarding learning-based methods~\cite{Fan2022DDPCCDD,Akhtar2022InterFrameCF}, since they are both being studied in the MPEG AI-PCC group following the CTC for training and testing~\cite{MPEG-EE-AIDPCC}, we directly cite their results reported in the latest standard ad-hoc summary for a fair comparison~\cite{akhtar2022prpposal,ddpcc2022prpposal}.

{\bf Comparison to G-PCC/V-PCC.} 
As shown in Table~\ref{table:bdrate} and Fig.~\ref{fig:rdcurves}, in lossless mode, the proposed method reaches an average 45\% gain over the G-PCC anchor, e.g., 0.404 bpp versus 0.837 bpp when testing 100 frames; while in lossy mode, our method provides $\approx$78\% BD-Rate improvement against the anchor V-PCC.

\textbf{Comparison to learned static PCGC.} 
We present our BD-Rate gains over state-of-the-art learning-based methods used for static PCGC~\cite{Wang2021MultiscalePC, Wang2021SparseTM}. As also in Table~\ref{table:bdrate}, compared with PCGCv2~\cite{Wang2021MultiscalePC} that only supports the lossy coding, the proposed method attains 52.3\%/51.3\% BD-Rate reduction. In lossless mode, we improve the SparsePCGC~\cite{Wang2021SparseTM} by around 11\% on average (0.404/0.395 bpp versus 0.454/0.449 bpp); while in lossy mode, the gain over SparsePCGC is even higher, $>$ 22\% on average.  Note that the proposed method is extended on top of the SparsePCGC by introducing inter conditional coding. The resultant BD-Rate gain further confirms the superiority of the use of multiscale temporal priors in dynamic PCGC. 

\textbf{Comparison to learned dynamic PCGC.}
Further, we compare the proposed method with learning-based  dynamic  PCGC methods~\cite{Fan2022DDPCCDD,Akhtar2022InterFrameCF} in Table~\ref{table:bdrate}. We only compare lossy mode performance because their  solutions only support lossy compression. 
As shown, the proposed method significantly outperforms existing methods with approximately {28\%} and {50\%} BD-Rate gains over Fan~\textit{et al.}~\cite{Fan2022DDPCCDD} and Akhtar~\textit{et al.}~\cite{Akhtar2022InterFrameCF} on average. 
{Our superior performance mainly attributes to: 1) we adopt a multi-stage SOPA in lossless phase, which is more efficient than the use of lossless G-PCC  in~\cite{Akhtar2022InterFrameCF,Fan2022DDPCCDD}; 2) in the lossy phase, inter residual compensation at a fixed scale limits the performance of~\cite{Akhtar2022InterFrameCF,Fan2022DDPCCDD}.} Note that even using the same lossless G-PCC in our method as in~\cite{Fan2022DDPCCDD,Akhtar2022InterFrameCF}, the BD-Rate gains are also mostly retained due to the use of inter conditional coding.

We also visualize corresponding R-D curves in Fig.~\ref{fig:rdcurves}. It shows that our method consistently performs better than other methods across a wider range of bitrates.
It is also observed that Fan~{\it et al.}~\cite{Fan2022DDPCCDD} focus on high bitrates and cannot reach at bitrates below 0.06 bpp, while Akhtar~{\it et al.}~\cite{Akhtar2022InterFrameCF} is mostly applicable to low bit rates but performs poorly at high bitrates. 
{This occurs mainly due to the fixed scale setting in their respective lossy phase, i.e., Fan~\textit{et al.}~\cite{Fan2022DDPCCDD} downscales 2 times and Akhtar~\textit{et al.}~\cite{Akhtar2022InterFrameCF} downscales 3 times, for lossy compression.}
By contrast, our method provides flexible scale adjustment (i.e. high/medium/low bitrates with adaptive $m$ e.g., $m\in{1, 2, 3}$), and multiscale inter conditional coding  through simple-yet-effective feature concatenation. These improvements not only enable the support of both lossless and lossy compression but also yield SOTA performance.

{\bf Complexity.}
We  collect the runtime by respectively running the G-PCC, SparsePCGC, and the proposed method in lossless coding, as shown in Table~\ref{table:time} for complexity evaluation. The runtime is tested on an Intel Xeon Silver 4210 CPU and an Nvidia GeForce RTX 2080 GPU, which is just used as the intuitive reference to have a general understanding of the computational complexity. As seen,  the proposed method presents faster encoding and decoding than G-PCC when using GPU acceleration. The runtime increase relative to the SparsePCGC-based intra coding is marginal.

%% file: figs/table.tex
\begin{table*}[htbp]\small
\caption{Compression performance comparison with other methods (tested on 100/32 frames following the MPEG CTC~\cite{MPEG-EE-AIDPCC})}
\label{table:bdrate}
\centering
\begin{tabular}{|c|ccc|ccccc|}
\hline
\multirow{2}{*}{\textbf{\begin{tabular}[c]{@{}c@{}}sequences\\ (100/32)\end{tabular}}} & \multicolumn{3}{c|}{\textbf{lossless (bpp)}}                                                                 & \multicolumn{5}{c|}{\textbf{lossy (BD-Rate Gain \%)}}                                                                                                                                             \\ \cline{2-9} 
                                    & \multicolumn{1}{c|}{\textbf{G-PCC}}       & \multicolumn{1}{c|}{\textbf{SparsePCGC}}  & \textbf{Ours}        & \multicolumn{1}{c|}{\textbf{SparsePCGC}}  & \multicolumn{1}{c|}{\textbf{Fan}~\cite{Fan2022DDPCCDD}}         & \multicolumn{1}{c|}{\textbf{Akhtar}~\cite{Akhtar2022InterFrameCF}}      & \multicolumn{1}{c|}{\textbf{PCGCv2}~\cite{Wang2021MultiscalePC}}      & \textbf{V-PCC}       \\ \hline
{player}                     & \multicolumn{1}{c|}{0.824/0.812}          & \multicolumn{1}{c|}{0.445/0.441}          & 0.400/0.388          & \multicolumn{1}{c|}{-27.7/-24.2}          & \multicolumn{1}{c|}{-28.3/-28.0}          & \multicolumn{1}{c|}{-49.8/-49.1}          & \multicolumn{1}{c|}{-53.0/-51.3}          & -78.6/-78.9          \\ \hline
{dancer}                     & \multicolumn{1}{c|}{0.854/0.849}          & \multicolumn{1}{c|}{0.461/0.460}          & 0.425/0.425          & \multicolumn{1}{c|}{-11.9/-14.0}          & \multicolumn{1}{c|}{-26.7/-28.4}          & \multicolumn{1}{c|}{-49.1/-50.0}          & \multicolumn{1}{c|}{-45.8/-47.4}          & -77.6/-79.2          \\ \hline
{model}                      & \multicolumn{1}{c|}{0.840/0.811}          & \multicolumn{1}{c|}{0.460/0.451}          & 0.404/0.388          & \multicolumn{1}{c|}{-28.9/-26.6}          & \multicolumn{1}{c|}{-31.4/-31.7}          & \multicolumn{1}{c|}{-50.2/-53.4}          & \multicolumn{1}{c|}{-55.2/-55.0}          & -76.8/-78.8          \\ \hline
{exercise}                   & \multicolumn{1}{c|}{0.829/0/819}          & \multicolumn{1}{c|}{0.448/0.442}          & 0.388/0.379          & \multicolumn{1}{c|}{-31.0/-25.3}          & \multicolumn{1}{c|}{-26.0/-23.2}          & \multicolumn{1}{c|}{-49.7/-47.7}          & \multicolumn{1}{c|}{-55.3/-51.5}          & -77.6/-77.2          \\ \hline
\textbf{average}                    & \multicolumn{1}{c|}{\textbf{0.837/0.823}} & \multicolumn{1}{c|}{\textbf{0.454/0.449}} & \textbf{0.404/0.395} & \multicolumn{1}{c|}{\textbf{-24.9/-22.5}} & \multicolumn{1}{c|}{\textbf{-28.1/-27.8}} & \multicolumn{1}{c|}{\textbf{-49.7/-50.0}} & \multicolumn{1}{c|}{\textbf{-52.3/-51.3}} & \textbf{-77.7/-78.5} \\ \hline
\end{tabular}
\end{table*}

\begin{table}[htbp]\small
\centering
\caption{Average runtime comparison in lossless mode}
\label{table:time}
\begin{tabular}{|c|c|c|c|}
\hline
\textbf{Time (s/frame)} & \textbf{G-PCC} & \textbf{SparsePCGC} & \textbf{Ours} \\ \hline
\textbf{Enc}      & 4.75            & \textbf{1.82}    & {1.96}  \\ \hline
\textbf{Dec}      & 2.60            & \textbf{1.66}    & {1.82}  \\ \hline
\end{tabular}
\end{table}

%% file: 6_conclusion.tex
\section{Conclusion}

 This paper presents the compression of dynamic point cloud geometry, which incorporates the multiscale temporal priors into the multiscale sparse representation framework to enable inter conditional coding  across temporal frames.
 Extensive experiments demonstrate that the proposed approach achieves SOTA performance in both lossy and lossless modes when compressing the dense object point cloud geometry.